\pdfoutput=1

\documentclass[journal]{IEEEtran}

\usepackage{graphicx}
\usepackage{xcolor}
\usepackage{array}
\usepackage{textcomp}
\usepackage[cmex10]{amsmath}
\usepackage{amssymb}
\usepackage{mathtools}
\usepackage{mathrsfs}
\usepackage{multirow}
\usepackage{url}
\usepackage{fixltx2e}
\usepackage{arydshln}
\usepackage{hyperref}
\usepackage[all]{hypcap}
\usepackage{pdfpages}

\newcommand{\IIGsources}{\url{https://github.com/CTU-IIG}}

\begin{document}

\title{Energy Optimization of Robotic Cells}

\author{Libor~Bukata, P\v{r}emysl \v{S}\r{u}cha, Zden\v{e}k Hanz\'{a}lek, and Pavel Burget
\thanks{Manuscript received October 08, 2015; revised March 07, 2016 and August 05, 2016; and accepted October 22, 2016}
\thanks{Libor Bukata and Zden\v{e}k Hanz\'{a}lek are with the Czech Institute of Informatics, Robotics,
and Cybernetics, Czech Technical University in Prague, Zikova 1903/4, 166\,36, Prague 6, Czech Republic.
(e-mail: bukatlib@fel.cvut.cz, hanzalek@fel.cvut.cz)}
\thanks{P\v{r}emysl \v{S}\r{u}cha and Pavel Burget are with the Faculty of Electrical Engineering, Department of Control Engineering,
Czech Technical University in Prague, Karlovo n\'{a}m\v{e}st\'{i} 13, 121\,35, Prague 2, Czech Republic.
(e-mail: suchap@fel.cvut.cz, burgetpa@fel.cvut.cz)}
\thanks{Copyright (c) 2016 IEEE. Personal use of this material is permitted. However, permission to use this material
for any other purposes must be obtained from the IEEE by sending a request to pubs-permissions@ieee.org.}
}

\markboth{IEEE TRANSACTIONS ON INDUSTRIAL INFORMATICS}
{Bukata {\textit{et al.}}: Energy Optimization of Robotic Cells}

\null
\includepdf[pages=1,fitpaper,noautoscale]{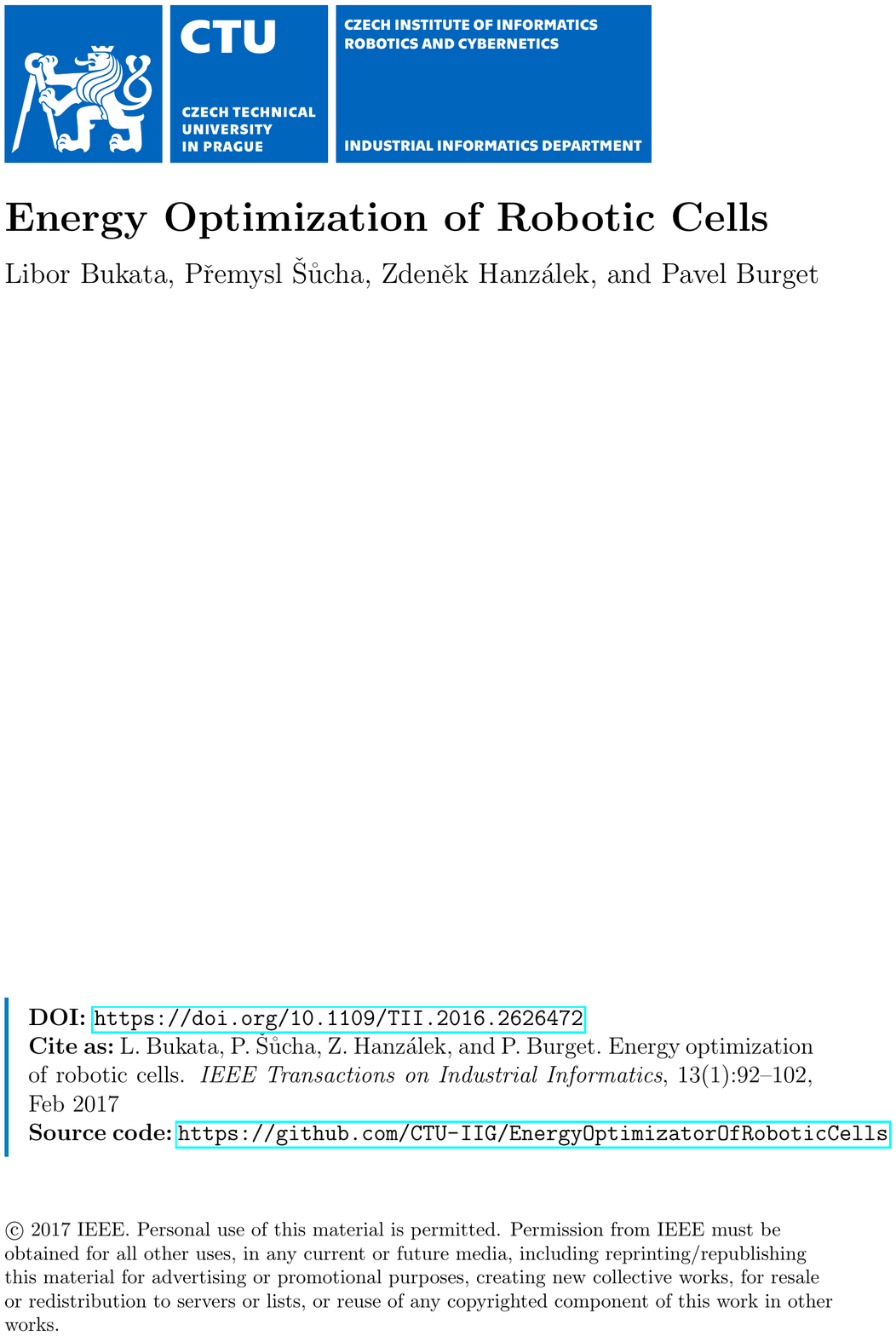}
\maketitle


\begin{abstract}


%


This study focuses on the energy optimization of industrial robotic cells, which is essential for sustainable production in the long term.
A holistic approach that considers a robotic cell as a whole toward minimizing energy consumption is proposed.
The mathematical model, which takes into account various robot speeds,
positions, power-saving modes, and alternative orders of operations,
can be transformed into a mixed-integer linear programming formulation that is, however, suitable only for small instances.
To optimize complex robotic cells, a hybrid heuristic accelerated by using multi-core processors
and the Gurobi simplex method for piecewise linear convex functions is implemented.
The experimental results showed that the heuristic solved 93\,\% of instances
with a solution quality close to a proven lower bound.
Moreover, compared with the existing works, which typically address problems with 3 to 4 robots,
this study solved real-size problem instances with up to 12 robots and considered more optimization aspects.
The proposed algorithms were also applied on an existing robotic cell in \v{S}koda Auto.
The outcomes, based on simulations and measurements, indicate that, compared with the previous state
(at maximal robot speeds and without deeper power-saving modes), the energy consumption
can be reduced by about 20\,\% merely by optimizing the robot speeds and applying power-saving modes.
All the software and generated data sets used in this research are publicly available.

\end{abstract}


\IEEEpeerreviewmaketitle


\section{Introduction}

\begin{figure}[htb]
	\centering
	\includegraphics[width=0.47\textwidth]{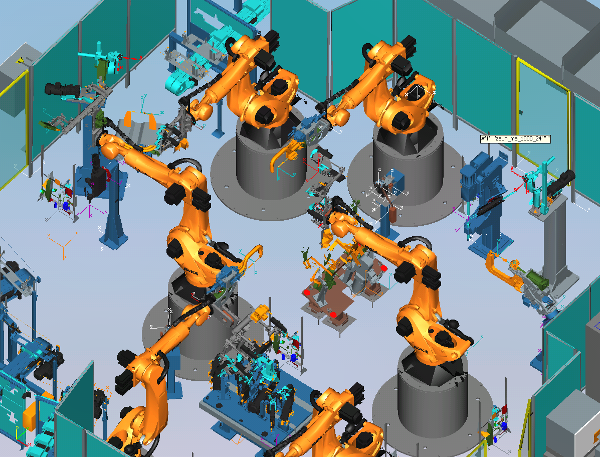}
	\caption{A typical robotic cell in the industry.}
	\label{fig:rob_cell}
\end{figure}

\IEEEPARstart{R}{obotic cells}, which are broadly used in heavy-duty industries, are highly complex systems (see Figure~\ref{fig:rob_cell}) that consist mainly of
industrial robots, conveyors, (stationary) welding/gluing guns, PLC controllers, and physical barriers (e.g., fencing).
Because the primary objective is to guarantee the repeatability of production, the robots carry out the same operations periodically;
i.e., they follow fixed cyclic event-based schedules, the execution of which can be conditioned by some signals, such as an acknowledgment of a free work space.
Signaling is also used for inter-robot synchronization to avoid collisions.
Thus far, robotic cells are often optimized with respect to their production rate (see, e.g., Dawande et al. \cite{Dawande2007} for a survey);
however, the pressure to reduce energy consumption in the industry is growing due to increasing costs of electricity,
the energy savings goal of the European Union \cite{EUplan}, and other initiatives and requirements of ecological groups.
As a result, there is a high demand for energy-efficient solutions that enable sustainable development in the long term.
One of the areas in which a significant amount of energy can be saved is the manufacturing process.
For example, industrial robots consume approximately 8\,\% of the energy needed during the production of cars (Meike and Ribickis \cite{EnEfUseOfRobotics}).

The present study addresses the energy optimization of robotic cells not only in the automotive industry.
A mathematical model of the robotic cell is presented, and optimization algorithms are proposed.
The primary goal of the optimization is to minimize the energy consumption of a robotic cell for a given production rate by
changing the robot \emph{speeds}, \emph{positions} (including the rotation of a gun),
and \emph{order} of operations and by applying the robot \emph{power-saving modes}.
These optimization aspects have been identified as important for energy consumption
based on measurements made on a small industrial KUKA robot.
Such optimization has significant potential for energy savings
because robotic cells are usually designed under time pressure with the main objective of meeting a desired production rate.
As a result, the speed of robot movements is typically maximal, although these mostly result in long idle times.
Such performance leads to a power profile that sharply fluctuates during production; for instance,
a regular 6-axis industrial robot with a 200\,kg payload requires between 0.5 and 20\,kW of power (Meike and Ribickis \cite{EnEfUseOfRobotics}).

For stationary robots, the robot brakes can slightly alleviate the energy burden if they are released early
because the motors do not consume energy in holding the robot in a certain position,
and the power needed to keep the brakes open does not apply to the released brakes.
On one hand, the frequent use of brakes could result in significant energy savings;
on the other hand, caution must be exercised because the number of brake cycles is limited to 5 million,
and the expected lifetime of robots is about 15 years (Meike et al. \cite{MeikeJour}).
Some robots, such as those from KUKA, also support the PROFIenergy profile \cite{PROFIenergy},
which allows controlling the power consumption by sending commands through a PROFINET network.
This enables deeper energy-saving modes for robots, such as bus power-off or hibernate.
Several experimental facilities have been using PROFIenergy, such as the Mercedes-Benz plant
in Sindelfingen, in which various experiments have been carried out \cite{Mercedes}.
Meike and Ribickis \cite{EnEfUseOfRobotics} provide more detailed information on energy-saving techniques.

The current research on the energy optimization of robotic cells often concentrates on the optimization of
individual robot trajectories (local optimization) with respect to the physical limitations of robots
and the obstacles to be avoided (e.g., \cite{TrajOpt1, TrajOpt2}, and \cite{TrajOpt3}).
However, to reach the full energy-saving potential, it is necessary to consider the robotic cell as a whole (global optimization).
Such energy optimization was proposed in a pilot work of Wigstr\"{o}m and Lennartson \cite{OptRob4},
in which a nonlinear mathematical model was formulated and solved by using various algorithms.
Nevertheless, the authors neglected the energy-saving modes and alternative positions of robots,
and their algorithms were benchmarked on an instance of job shop scheduling problem.

In the work of Mashaei and Lennartson \cite{OptRob5}, an energy model of a pallet-constrained flow shop problem was formulated
to find an optimal switching control strategy for achieving the desired throughput and minimal energy consumption.
Idle states of machines were also taken into account to reduce the energy consumption when the machine was not working.
However, the model required a line with a particular structure, i.e., a closed-loop pallet system;
therefore, it was not generally applicable to robotic cells.

A special structure, i.e., a rotationally arranged robotic cell, was also considered by Foumani et al. \cite{IEEE-II1},
who proposed a methodology for maximizing the production rate subject to timing constraints, in which
multifunction robots that performed an extra operation during the movement were taken into account to reduce the cycle time.
For example, the robot could measure the dimensions of a workpiece during its transportation by a Grip-Gage-Go gripper.
Sindi\v{c}i\'{c} et al. \cite{IEEE-II2} proposed a method that determined
whether a flexible manufacturing system was stable for given repeatable sequences.
The proposed method used a machine-job incidence matrix that was typically smaller
than matrices describing the structure of Petri nets, thus reducing the complexity of the analysis.
The authors applied their approach on a system with three robots and four machines.

On the boundary of local and global optimization is the work of Meike et al. \cite{MeikeJour, MeikeConf},
in which the last movement to the robot home position and the subsequent waiting period were optimized to save energy.
Compared with the work of Wigstr\"{o}m and Lennartson \cite{OptRob4}, in this study the robot brakes were considered at the robot home position.
Although only a relatively small part of the robotic cell was considered,
the authors estimated that the energy savings for the robotic cell reached 7.3\,\%.

To push the aforementioned research further, the present work proposes a \emph{holistic mathematical model} for the energy optimization of robotic cells.
In contrast to the existing works \cite{OptRob4, OptRob7, OptRob1}, this model supports the energy-saving modes and alternative positions of robots.
Because the continuous manufacturing process is considered and not the production-free time, only robot brakes and bus power-off
energy-saving modes are applicable in practice due to the long transition time of deeper energy-saving modes.
In the proposed model, the \emph{robot cycle time}, which is known as the period from cyclic scheduling, corresponds to
the time interval (\texttildelow 1/throughput) between two outgoing workpieces after the start-up phase.
The robot cycle time is usually determined according to the desired production rate of a product.
The duration of the start-up phase, which is called the lead time in cyclic scheduling, is the total
time required for one (first) workpiece to be processed by the robotic cell; or in other words,
it is the time difference between the entering and the leaving time of a given workpiece.
The cycle time is typically shorter than the duration of the start-up phase; thus,
there are usually more unfinished workpieces in the robotic cell at a given time.
Such parallelism is analogous with the cyclic scheduling
of the processor pipeline (see, e.g., \cite{cyclicSchedulingPipeline1} and \cite{cyclicSchedulingPipeline2}).
However, Wigstr\"{o}m and Lennartson \cite{OptRob4} presumed that the whole robotic cell
is dedicated to one workpiece during the entire processing time, which the authors called the work cycle time;
hence, the overall throughput may be limited because there is no parallelism.
This different view of time distinguishes the problem in the present work.
Furthermore, the proposed model is suitable for both existing and new robotic cells because the number of optimized aspects,
such as the selection of robot positions, speeds, and energy-saving modes, can be adjusted.

Based on the model, the \emph{mixed-integer linear programming} (MILP) problem
is formulated and subsequently solved by applying the Gurobi solver.
The experimental results, however, show that only small problems are solvable close to optimality
and that bigger problems are intractable for the state-of-the-art MILP solvers.
Therefore, this research proposes a \emph{parallel hybrid heuristic} that solves some bigger instances in a short time.
The proposed heuristic is very efficient and fast, thanks to the application of a special Gurobi simplex algorithm for piecewise linear convex functions
and the parallelization that enables the algorithm to exploit the power of all processor cores.
The proposed algorithms are tested on a real robotic cell from \v{S}koda Auto (see Section \ref{sec:casestudy}) and generated data sets.
The estimation, based on simulations and measurements of power profiles in \v{S}koda Auto, indicate that up to
17.6\,\% of energy can be saved merely by changing the speed of robot movements
and that even more savings can be achieved if energy-saving modes are applied.
These results are partially supported by Paryanto et al. \cite{MeasTraj}, who identified
the speed of robot movements as crucial for energy efficiency.
In addition, a \emph{generator} of problem instances is implemented
to enable other researchers to evaluate their algorithms given the lack of instances that are publicly available.
Both the algorithms and the generator are published as \emph{open-source software} and can be downloaded from \IIGsources.

The main contributions of the present work are the extended mathematical model, which considers the power-saving modes and
alternative positions of robots, and the design of the parallel hybrid heuristic,
which has been shown to be effective in the experiments.
The results of the case study from \v{S}koda Auto indicate that the proposed approach
may considerably reduce the energy consumption of robotic cells.
The rest of the report is structured as follows.
Section~\ref{sec:prob_st} sets out the formal specification of the problem,
Section~\ref{sec:ilp} presents the MILP model,
and Section~\ref{sec:heur} introduces the parallel heuristic algorithm for solving the problem.
The performance of the proposed algorithms is tested on benchmark instances in Section \ref{sec:res},
and the case study from \v{S}koda Auto is discussed in Section~\ref{sec:casestudy}.
Finally, the conclusions are presented in Section~\ref{sec:conclusions}.

\section{Problem Statement}\label{sec:prob_st}


The energy optimization problem of the robotic cell can be defined as follows.
There is a set of robots $\mathcal{R} = \{r_1, \dots, r_{\left|\mathcal{R}\right|}\}$
and their graphs denoted as $G_r = (V_r, E_r)$ where nodes $V_r$ are static activities (e.g., waiting or welding),
and edges $E_r$ are dynamic activities, which define the possible robot moves between static activities.
Let $V = \bigcup_{\forall r \in \mathcal{R}} V_r$, $E = \bigcup_{\forall r \in \mathcal{R}} E_r$, and $A_r = V_r \cup E_r$.

Each static activity $v \in V$ has the assigned set $L_v$ of possible robot positions, i.e., locations,
in which the motionless robot either works or waits for a signal.
During this stationary phase lasting from $\underline{d}_v$ to $\overline{d}_v$,
one of the robot power saving modes $m \in M_r$, including a dummy power-saving mode
for the robot held by motors, can be applied if the activity duration $d_v \geq \underline{d}^m$,
where $\underline{d}^m$ is the minimal time to switch the mode on.
The energy consumption of activity $v$ is then $W_v = p_{v,l}^m d_v$,
where $p_{v,l}^m$ is the robot input power at location $l \in L_v$ for mode $m \in M_r$.

The dynamic activity, i.e., edge $e \in E_r$, consists of the set of trajectories $T_e$ between $v_1$ and $v_2 \in V_r$,
where each trajectory $t \in T_e$ interconnects two locations, and its duration is from $\underline{d}_e^t$ to $\overline{d}_e^t$.
The energy consumption of activity $e \in E_r$ is the function of the selected trajectory $t \in T_e$,
the duration of the movement $d_e$, and a convex function $f^t_e$ mapping $d_e$ to energy consumption $W_e$.
The form of the function (see Vergnano et al. \cite{OptRob6}, p. 389, Eq. 12)
is $f^t_e(d_e) = \sum_{i=1}^{5} C^t_{e,i} d_e^{2-i}$, where $C^t_{e,i}$ are constants.

Let $A = V \cup E$ be the set of all activities; then, $S_a = \mathrm{suc}(a)$
and $P_a = \mathrm{pred}(a)$ are the set of successors and predecessors of activity $a \in A$, respectively.
If $|S_v| > 1$ for $v \in V$, then the dynamic activities $e \in S_v$ are optional
because only one trajectory $t \in T_e$ will be selected as the leaving one.
The optional activities, denoted as $E_{\mathcal{O}} \subseteq E$, can model alternative orders of operations.
Note that $|S_e| = 1$ for $\forall e \in E$.

The closed path of robot $r \in \mathcal{R}$, including the order of operations,
can be represented as a directed Hamiltonian circuit, $\mathcal{HC}^{\mathrm{loc}}_r = (l_1, \dots, l_{|V_r|})$,
through the activity locations, where each $v \in V_r$ is visited only once, with the exception of a closing activity.
The order of operations (activities), regardless of the selected locations, can be expressed in the form of a directed Hamiltonian circuit,
$\mathcal{HC}^{\mathrm{act}}_r = (v_1, \dots, v_{|V_r|})$, where the last activity $v_{|V_r|}$, denoted as $v^h_r$, closes the cycle of robot $r$.
The last activity $v^h_r$, regardless of the work of robot $r$, is predetermined as
\emph{home positions} of robot $r$, where the robot typically waits for its next cycle.
Note that $\mathcal{HC}^{\mathrm{act}}_r$ can be derived from $\mathcal{HC}^{\mathrm{loc}}_r$; however, the converse is not true.
The robot cycle time $\mathrm{CT}$, i.e., the production cycle time of the robotic cell, has to be fulfilled; in other words,
$\sum_{\forall a \in A_r^*} d_a = \mathrm{CT}$ for $\forall r \in \mathcal{R}$, where $A_r^* \subseteq A_r$ refers to all activities
on the related Hamiltonian circuit $\mathcal{HC}^{\mathrm{act}}_r$.

To guarantee that the robots cooperate with each other at the right time and location, there are
global constraints for \emph{time synchronization} and \emph{spatial compatibility}.
Correct timing is ensured by inter-robot time lags $E_{\mathcal{T\hspace{-0.5mm}L}}$,
where each time lag $(a_1, a_2) \in E_{\mathcal{T\hspace{-0.5mm}L}}$
has the fixed length $l_{a_1, a_2} \in \mathbb{R}$ and the height $h_{a_1, a_2} \in \mathbb{Z}$.
The length $l_{a_1, a_2}$ denotes a time shift between $a_1 \in A_{r_i}$ and $a_2 \in A_{r_j}$,
and the height $h_{a_1, a_2}$ enables addressing the current and previous robot cycles.
Both terms, which are well-known in cyclic scheduling, define the time relation as follows:
$s_{a_2} \geq s_{a_1}+l_{a_1, a_2}-\mathrm{CT} h_{a_1, a_2}$,
where $s_{a_1}$ and $s_{a_2}$ are the start times of activities $a_1$ and $a_2$, respectively.

To ensure that the workpiece is passed correctly from one robot to another, i.e., that
the workpiece is picked up from the same place where it was put, it is necessary to define the spatial compatibility as follows.
Let $v_1$ and $v_2 \in V$ be two static activities in which an inter-robot handover is carried out;
then, the set of compatible location pairs is $Q_{v_1 \times v_2} \subseteq L_{v_1} \times L_{v_2}$, and
set $Q_{l_i} = \{\forall l_j : (l_i, l_j) \in Q_{v_1 \times v_2}\;\text{or}\;(l_j, l_i) \in Q_{v_1 \times v_2}\}$
contains all the locations (robot positions) compatible with location $l_i$.
Finally, because the concurrent work of robots may result in collisions,
time disjunctive combinations of trajectories and locations are specified as quadruplets $(a_1, g_1, a_2, g_2) \in C$,
where $g_i \in L_{a_i}$ if $a_i \in V$; otherwise, $g_i \in T_{a_i}$.

The goal of the optimization is to find the closed paths of robots (i.e., $\mathcal{HC}^{\mathrm{loc}}_r$),
determine the activity timing (i.e., the assignment of $s_a$ and $d_a$ for $\forall a \in A$),
and decide which power-saving mode of the robot to apply for each static activity $v \in V$ such that global constraints are satisfied,
collisions are prevented, and the energy consumption (i.e., $\sum_{\forall a \in A} W_a$) is minimized.

\subsection{Example}

\begin{figure}
	\centering
	\includegraphics[width=0.5\textwidth]{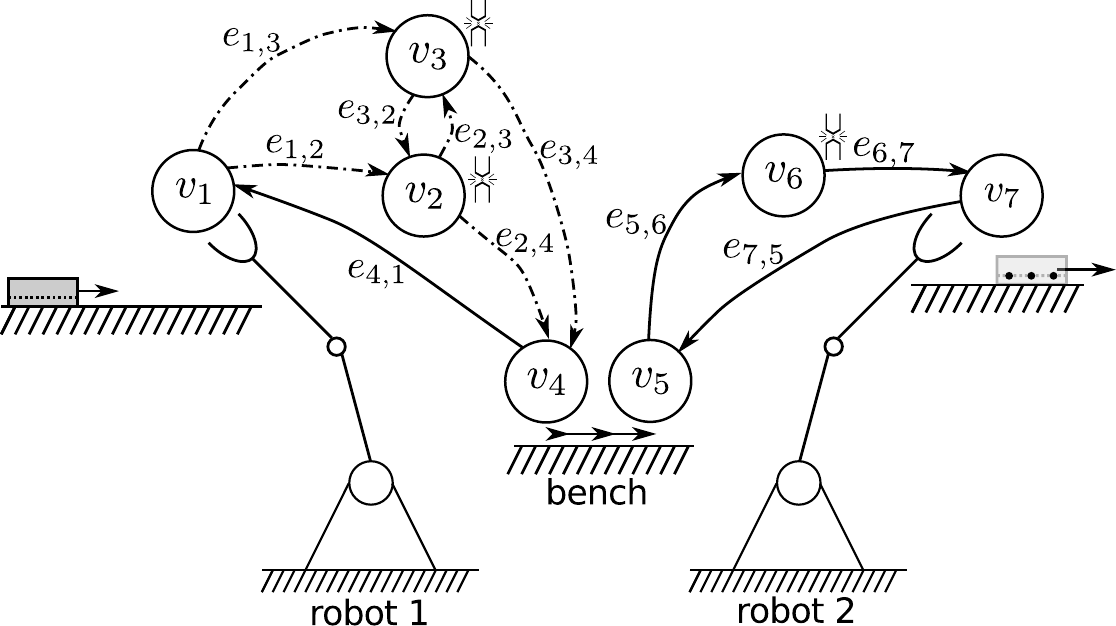}
	\label{fig:psexample}
	\caption{Robotic cell with two robots that perform welding operations.}
\end{figure}

Figure~\ref{fig:psexample} shows a robotic cell consisting of two robots: $r_1$ and $r_2 \in \mathcal{R}$.
Robot $r_1$ takes the weldment, carries out two spot-welding operations, and subsequently puts the weldment on a bench.
Robot $r_2$ takes the weldment from the bench, carries out an additional spot-welding operation, and puts the weldment on a conveyor.
The process flow of each robot can be expressed as a graph $G_r$; such as $G_{r_1} = (V_{r_1}, E_{r_1})$, where
$V_{r_1} = \{v_1, v_2, v_3, v_4\}$ and $E_{r_1} = \{e_{1,2}, e_{1,3}, e_{2,3}, e_{2,4}, e_{3,2}, e_{3,4}, e_{4,1}\}$.
Nodes $v_2, v_3, v_6$ correspond to spot-welding operations, whereas $v_1, v_5$ and $v_4, v_7$ are take and put operations, respectively.
The edges define how to move from one operation to another, and the edge style determines whether
the move is mandatory (solid line), or optional (dashed line).
The dashed edges, i.e., the optional dynamic activities $E_\mathcal{O}$, enable modeling
alternative orders of operations; in this case, there are two possible orders: $(v_1, v_2, v_3, v_4)$ and $(v_1, v_3, v_2, v_4)$.
The graphs shown in Figure~\ref{fig:psexample} have activity-level granularity
because each edge is a dynamic activity and each node is a static activity.
However, a lower-level of granularity is achievable because each node $v$
has locations $L_v$, and each edge $e$ contains trajectories $T_e$.
Because the weldment has to be passed at the right place at the right time,
spatial and time synchronization between robots is necessary.
For example, if $L_{v_4} = \{l_{4,1}, l_{4,2}\}$ and $L_{v_5} = \{l_{5,1}, l_{5,2}\}$, i.e., $v_4$ and $v_5$ have 2 locations each, then
the spatial compatibility can be defined as $Q_{l_{4,1}} = \{l_{5,2}\}$ and $Q_{l_{4,2}} = \{l_{5,1}\}$.
To ensure correct timing, two time lags need to be added:
$s_{v_5} \geq s_{e_{4,1}}+l_{e_{4,1}, v_5}$ and $s_{v_4} \geq s_{e_{5,6}}+l_{e_{5,6}, v_4} - \mathrm{CT}$.
The first time lag ensures that robot $r_2$ takes the weldment from the bench after $r_1$ has put it there.
The second time lag guarantees that robot $r_2$ has left the bench before $r_1$ puts another weldment there.
The lengths of the time lags are set to the required time for a robot to safely leave the bench.

\section{Mixed-Integer Linear Programming Model}\label{sec:ilp}


The problem is intrinsically nonlinear due to $f_e^t$ being convex functions;
nevertheless, a MILP model can be formulated if the functions are piecewise linearized.
The piecewise linear functions $\hat{f}_e^t$ are propagated through constraints \eqref{ef:dynamic} to the criterion \eqref{ilp:criterion},
where $\overline{W}$ is an upper bound on energy consumption, $y_e^t$ is a binary variable
that determines whether or not the trajectory $t \in T_e$ is selected for activity $e \in E$, $B$ is a set of indices,
and $k_{e,b}^t$, $q_{e,b}^t$ are coefficients of the $b$-th linear function approximating $f_e^t$.
The number of segments $|B|$ of each $\hat{f}_e^t$ can be adjusted to meet the desired accuracy of the approximation.
There is no need to introduce binary variables for each segment of the function
because functions $f_e^t$ are convex, and the criterion ensures that the right linear function is active.
The energy consumption of static activities is calculated by constraints~\eqref{ef:static},
where $x_v^l$ determines whether or not location $l$ is selected for $v \in V$,
and $z_v^m$ is set to true if and only if mode $m \in M_r$ is applied for activity $v \in V$.

\begin{alignat}{1}
	\text{Min} & \sum_{\forall a \in A} W_a \label{ilp:criterion} \\[1mm]
s.t.	& \quad p_{v,l}^m d_v - \overline{W} \left(2-z_v^m-x_v^l\right) \leq W_v \label{ef:static} \\
	& \qquad\quad \forall r \in \mathcal{R}, \forall v \in V_r, \forall l \in L_v, \forall m \in M_r \notag \\
	& \quad k_{e,b}^t d_e + q_{e,b}^t - \overline{W} \left(1-y_e^t\right) \leq W_e \label{ef:dynamic} \\
	& \qquad\quad \forall e \in E, \forall t \in T_e, \forall b \in B \notag
\end{alignat}

Assignment constraints \eqref{onepoint}, \eqref{onemode}, and \eqref{onetraj} ensure
the correct selection of locations, power saving modes, and trajectories for activities.
Note that optional activities $E_\mathcal{O}$ are omitted in constraints \eqref{onetraj} because they may or may not be carried out.
If an optional activity $e \in E_\mathcal{O}$ is not performed, then none of its trajectories is selected.
Therefore, $W_e$ is zero because all the constraints \eqref{ef:dynamic} for activity $e$
are deactivated, and criterion \eqref{ilp:criterion} pushes $W_e$ down to zero due to the minimization.

\begin{alignat}{1}
	\sum_{\forall l \in L_v} x_v^l = 1 & \qquad \forall v \in V \label{onepoint} \\
	\sum_{\forall m \in M_r} z_v^m = 1 & \qquad \forall r \in \mathcal{R}, \forall v \in V_r \label{onemode} \\
	\sum_{\forall t \in T_e} y_e^t = 1 & \qquad \forall e \in E \setminus E_{\mathcal{O}} \label{onetraj}
\end{alignat}

Flow constraints \eqref{flow1} and \eqref{flow2} state that
if the robot moves to location $l$ then it also has to move away from the same location.

\begin{alignat}{1}
	\sum_{\forall e \in P_v} \sum_{\forall t=(l_\textrm{from}, l) \in T_e} y_e^t = x_v^l & \qquad \forall v \in V, \forall l \in L_v \label{flow1} \\
	\sum_{\forall e \in S_v} \sum_{\forall t=(l, l_{\textrm{to}}) \in T_e} y_e^t = x_v^l & \qquad \forall v \in V, \forall l \in L_v \label{flow2}
\end{alignat}

Constraints \eqref{prec} to \eqref{pathsel2} enforce the time ordering of activities.
Some precedences of activities are mandatory (see constraints \eqref{prec} and \eqref{precend}) due to the structure of graph $G_r(V_r,E_r)$,
whereas others are optional (see constraints \eqref{pathsel} to \eqref{pathsel2}) because some $e \in E_{\mathcal{O}}$ do not have to be carried out.
The binary variables $w_{e,v}$ determine whether or not activity $e \in E_{\mathcal{O}}$ is performed and which order of operations is selected.

\begin{alignat}{1}
	& s_{a_2} = s_{a_1}+d_{a_1} \label{prec} \\
	& \qquad \forall a_1 \in A \setminus E_{\mathcal{O}}, \forall a_2 \in S_{a_1}, \nexists v_r^h = a_1 \notag \\
	& s_{e} = s_{v_r^h} + d_{v_r^h}-\mathrm{CT} \label{precend} \\
	& \qquad \forall v_r^h \in V, \forall e \in S_{v_r^h} \notag \\
	& \sum_{\forall t \in T_e} y_e^t = w_{e,\mathrm{suc}(e)} \qquad \forall e \in E_{\mathcal{O}} \label{pathsel} \\
	& s_v +  \left(1 - w_{e,v}\right) \mathrm{CT} \geq s_e+d_e \label{pathsel1} \\
	& \qquad \forall e \in E_{\mathcal{O}}, \forall v \in S_e \notag \\
	& s_v - \left(1 - w_{e,v}\right) \mathrm{CT} \leq s_e+d_e \label{pathsel2} \\
	& \qquad \forall e \in E_{\mathcal{O}}, \forall v \in S_e \notag
\end{alignat}

Restrictions on the duration of static and dynamic activities are reflected in constraints \eqref{durstat} and \eqref{durdyn}, respectively.

\begin{alignat}{1}
	& \max (\underline{d}_v, \underline{d}^m) z_v^m \leq d_v \leq \overline{d}_v \label{durstat} \\
	& \qquad \forall r \in \mathcal{R}, \forall v \in V_r, \forall m \in M_r \notag \\
	& \underline{d}_e^t y_e^t \leq d_e \leq \overline{d}_e^t + \mathrm{CT} \left(1-y_e^t\right) \label{durdyn} \\
	& \qquad \forall e \in E, \forall t \in T_e \notag
\end{alignat}

The problem described by Equations \eqref{ilp:criterion} to \eqref{durdyn} is robot independent;
i.e., the constraint matrix of the MILP formulation is block diagonal; thus, each robot can be treated separately.
In the following constraints, however, the robots are coupled together
with regard to time synchronization and spatial compatibility.

\begin{alignat}{1}
	& s_{a_2}-s_{a_1} \geq l_{a_1, a_2} - \mathrm{CT} h_{a_1, a_2} \qquad \forall (a_1, a_2) \in E_{\mathcal{T\hspace{-0.5mm}L}} \label{timelags} \\
	& x_{v_1}^{l_1} \leq \sum_{\forall l_2 \in L_{v_2}, l_2 \in \mathrm{Q}_{l_1}} x_{v_2}^{l_2} \label{sptcomp} \\
	& \qquad \forall v_1 \in V, \forall l_1 \in L_{v_1}, \left|Q_{l_1}\right| > 0 \notag
\end{alignat}

Collisions between robots are resolved by constraints \eqref{collisions1} and \eqref{collisions2}
where $u_{a_i}^{g_i}$ is either $x_{a_i}^{g_i}$ or $y_{a_i}^{g_i}$ depending on whether $a_i \in V$ or $a_i \in E$,
and the binary variables $c_{o}^n$ determine the execution order of $a_i$ and $a_j$
for the multiples of cycle time $\mathrm{CT}$ if the collision applies.
As an example, consider $x_{a_1}^{g_1}$ and $y_{a_2}^{g_2}$ to be a colliding pair for a multiple of cycle time $n = 2$.
If location $g_1$ and movement $g_2$ are selected, i.e., if the variables $u_{a_1}^{g_1}$ and $u_{a_2}^{g_2}$ are equal to one,
then exactly one of constraints \eqref{collisions1} and \eqref{collisions2} is active for this collision pair and multiple $n = 2$;
therefore, either $s_{a_2}+2\mathrm{CT} \geq s_{a_1}+d_{a_1}$ or $s_{a_1} \geq s_{a_2}+d_{a_2}+2\mathrm{CT}$ has to be satisfied.
In case the collision does not apply, i.e., if location $g_1$ or movement $g_2$ is not selected, then these two equations remain inactive.

\begin{alignat}{1}
	& s_{a_2} + n \mathrm{CT} + 2\left|\mathcal{R}\right| \mathrm{CT} (3-c_{o}^{n}-u_{a_1}^{g_1}-u_{a_2}^{g_2}) \geq s_{a_1}+d_{a_1} \notag \label{collisions1} \\
	& \quad \forall o=(a_1, g_1, a_2, g_2) \in C, \forall n \in \{-\left|\mathcal{R}\right|, \dots, \left|\mathcal{R}\right|\} \\
	& s_{a_1} + 2 \left|\mathcal{R}\right| \mathrm{CT} (2+c_{o}^n-u_{a_1}^{g_1}-u_{a_2}^{g_2}) \geq s_{a_2}+d_{a_2}+ n \mathrm{CT} \notag \label{collisions2} \\
	& \quad \forall o=(a_1, g_1, a_2, g_2) \in C, \forall n \in \{-\left|\mathcal{R}\right|, \dots, \left|\mathcal{R}\right|\}
\end{alignat}

All the variables of the model are either nonnegative floats or binary variables, as expressed in Equation~\eqref{domain}.

\begin{alignat}{1}
	& W_a, s_a, d_a \in \mathbb{R}_{\ge 0} \qquad x_v^p, z_v^m, y_e^t, c_{o}^n, w_{e,v} \in \mathbb{B} \label{domain}
\end{alignat}

\section{Parallel Heuristic Algorithm}\label{sec:heur}




Motivated by the inability of MILP solvers to solve bigger instances, this study proposes
a parallel hybrid heuristic based on a linear programming (LP) solver
that iteratively solves partially fixed problems, called \emph{tuples},
for selected locations, power-saving modes, and alternative orders.
Although the application of an LP solver to partially fixed problems
has been previously studied (see, e.g., \cite{hybridHeur1, hybridHeur2}, and \cite{hybridHeur3}),
the present work accelerates the heuristic by using a tailor-made Gurobi simplex algorithm
for piecewise linear convex functions and the parallelization.
The justification for the parallelization of combinatorial problems can be found in \cite{parallelAlg1, parallelAlg2}, and \cite{parallelAlg3}.

\begin{figure*}[!t]
	\centering
	\includegraphics[trim=0 5 15 25,clip,width=0.85\textwidth]{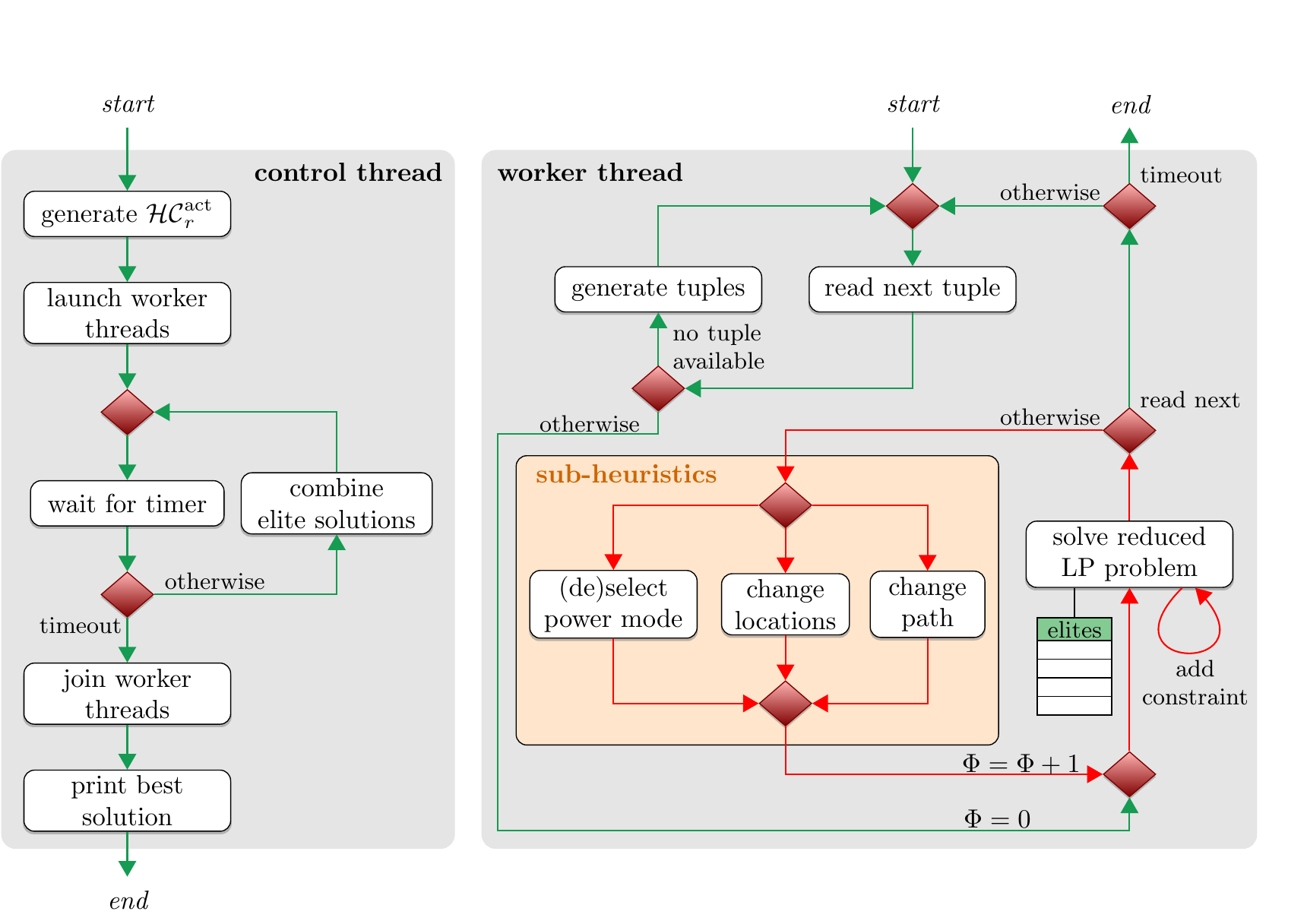}
	\label{fig:heur}
	\caption{Flowchart of the parallel hybrid heuristic.}
\end{figure*}

The flowchart of the heuristic, as shown in Figure~\ref{fig:heur}, can be divided
into two parts: the \emph{control thread} and \emph{worker threads}.
The control thread generates various alternative orders in the form of $\mathcal{HC}_r^{\mathrm{act}}$,
launches worker threads, and occasionally combines elite solutions to create promising tuples.
After the specified time limit $t_{\mathrm{max}}$, the control thread joins
the worker threads and prints the best solution if found.
The process flow of the worker thread consists of reading and generating tuples and then optimizing them iteratively
by alternating between solving a reduced LP problem and carrying out three sub-heuristics.
In general, each sub-heuristic performs a small modification of the tuple that may
reduce the energy consumption, and an LP solver evaluates a real impact of this modification.
The iterative optimization is stopped if there is no significant improvement
after $\Phi_{\mathrm{max}}$ iterations, in which case the next tuple is read,
or if the time limit $t_{\mathrm{max}}$ is exceeded, in which case the thread is terminated.

The heuristic (see Figure~\ref{fig:heur}) is accelerated by using multiple
worker threads that independently process tuples one by one from a list.
If the list becomes empty, then new tuples are added to it, and the process continues.
Because the tuples, alternative orders $\mathcal{HC}_r^{\mathrm{act}}$, and elite solutions are accessible
from all the threads, it is necessary to use synchronization primitives to guarantee consistent reads and writes.
Fortunately, the introduced overhead is negligible (see the experiments in Section~\ref{sec:res})
because the access time is very short compared with the time needed to solve the reduced LP problem.
The key parts of the heuristic are detailed in the following subsections.

\subsection{Generation of Alternatives}\label{sec:alt}



The order of operations of robot $r$, i.e., an alternative, is encoded as a Hamiltonian circuit $\mathcal{HC}_r^{\mathrm{act}}$ in graph $G_r(V_r,E_r)$.
Because there can be more alternative orders of operations (i.e., circuits), it is necessary to find some of them.
For this purpose, the finding of Hamiltonian circuits in graph $G_r(V_r, E_r)$
is transformed into a search for Hamiltonian paths in graph $G_r'(V_r', E_r')$ as follows.
Nodes $V_r' = V_r \setminus v_r^h \cup \{v_{\rightarrow}, v_{\leftarrow}\}$,
where $v_{\rightarrow}, v_{\leftarrow}$ are the starting and ending nodes, respectively.
The edges leaving $v_r^h$ are modified such that they start from $v_{\rightarrow}$; similarly
the edges entering $v_r^h$ are changed such that they end in $v_{\leftarrow}$.
Both the nodes and edges of $G_r'$ are weighted by the minimum possible duration $\underline{d}_a$ of the related activity;
i.e., $\min_{\forall t \in T_e} \underline{d}_e^t$ for $\forall e \in E_r'$
and $\max (\underline{d}_v, \min_{\forall m \in M_r} \underline{d}^m)$ for $\forall v \in V_r'$.
The goal is then to find random Hamiltonian paths from $v_{\rightarrow}$ to $v_{\leftarrow}$
such that $\sum_{\forall a \in \text{path}} \underline{d}_a \leq \mathrm{CT}$.
To ensure effective pruning during an iterative random tree search from node $v_{\rightarrow}$,
the obviously infeasible or completely searched sub-paths from $v_{\rightarrow}$ are detected and skipped.
The sub-path $(v_{\rightarrow}, \dots, a_i)$ is infeasible if there is an unvisited node $a_u$ that cannot be reached
or $|(v_{\rightarrow}, \dots, a_i)|+U(a_i, a_u)+U(a_u, v_{\leftarrow}) > \mathrm{CT}$,
where the first term is the sub-path length, and $U(a_i, a_j)$  is the length of the shortest path
from $a_i$ to $a_j$ calculated by the Floyd-Warshall algorithm.
If a Hamiltonian path is found, its fastest sequence through the activity locations
can be obtained by applying a shortest path algorithm for directed acyclic graphs.
If the duration of this sequence is longer than the robot cycle time, then the related Hamiltonian path
is neglected because it cannot lead to a feasible solution for the original problem.
Note that the Hamiltonian path and its fastest sequence through the locations can be easily
transformed back to $\mathcal{HC}^\mathrm{act}_r$ and $\mathcal{HC}^\mathrm{loc}_r$, respectively.
Because the algorithm is exact, i.e., it can prove the nonexistence of a Hamiltonian circuit of a given length,
it is possible to detect the infeasibility of some instances.
Moreover, the generation of alternatives is robot independent; therefore, the work is distributed
among threads to accelerate the initialization of the heuristic.

\subsection{Generation of Tuples}\label{sec:tupgen}



The generation of tuples, which are formally defined below, is crucial for finding initial feasible solutions;
therefore, the emphasis is placed on feasibility rather than on energy optimality during the generation.
At the beginning of the generation, a random alternative $\mathcal{HC}^{act}_r$
and its fastest closed path $\mathcal{HC}^{loc}_r$ through the locations (see Section~\ref{sec:alt}) are assigned to each robot $r$.
Only the fastest power-saving mode of robot $r$ is considered.
The closed paths are subsequently modified to meet the spatial compatibility as follows.
For each pair of static activities with the violated spatial compatibility
one fixing pair $q \in Q$ is selected with respect to the prolongation of the related closed paths;
i.e., the paths with a minimal duration close to the robot cycle time are penalized.
After the spatial compatibility is fixed, the tuple is created as a triple $\mathcal{T} = (\mathscr{A}, \mathscr{P}, \alpha: V \rightarrow M)$,
where $\mathscr{A}, \mathscr{P}$ are the selected alternatives and their closed paths, respectively,
and function $\alpha$ maps each activity $v \in V$ to its power mode $m \in M$, where $M = \cup_{\forall r \in \mathcal{R}} M_r$.
The process of tuple generation is applicable to two blocks shown in Figure \ref{fig:heur}: \emph{generate tuples} and \emph{combine elite solutions}.
However, in the second block, only the alternatives $\mathcal{HC}^{act}_r$ used in elite solutions are considered to intensify the search process.

\subsection{Reduced Linear Programming Problem}\label{sec:redlp}



The timing of a partial problem, i.e. a tuple, is determined by the reduced LP problem.
If the resulting solution is feasible, then
it is feasible for the original problem and can be added to the list of elite solutions
if it ranks among the top solutions in terms of energy consumption.
Otherwise, the solution is infeasible, and one of two cases arises:
the related tuple is not modified by any sub-heuristics, in which case another tuple is read;
or the previous tuple resulting in a feasible solution is loaded, and the next sub-heuristic is selected.

Before formally describing the reduced LP problem, it is necessary to define some sets extracted from tuple $\mathcal{T}$.
Set $E_\mathcal{T}$ contains all the dynamic activities in $\forall \mathcal{HC}_r^{\mathrm{act}} \in \mathscr{A}$,
where each dynamic activity $e \in E_\mathcal{T}$ has to be carried out (is mandatory) and has assigned its fixed trajectory $t$ as a pair $(e, t) \in \mathcal{F}_{\!\mathcal{T}}^1$;
similarly, each static activity $v \in V$ is linked to its location $l$ and power mode $m$ as a triple $(v, l, m) \in \mathcal{F}_{\!\mathcal{T}}^2$.
Then, the reduced LP problem can be stated as follows.
Criterion $\eqref{lp:crit}$ minimizes the sum of the piecewise linear convex functions $\hat{f}_e^t$
passed to the Gurobi LP solver as a list of function points or alternatively expressed
similarly to constraints \eqref{ef:static} and \eqref{ef:dynamic} for other solvers.
The equations marked with an asterisk are the same as the original ones (without the asterisk);
however the sets from which the constraints are generated are different.
Constraints \eqref{prec}*, \eqref{precend}*, and \eqref{timelags}*
establish precedences between activities $A_\mathcal{T} = V \cup E_\mathcal{T}$,
and constraints \eqref{durstat}* and \eqref{durdyn}* transform into the domain specification of $d_a$ variables.

\begin{alignat}{1}
	& \text{Min} \displaystyle \sum_{\forall (e, t) \in \mathcal{F}_{\!\mathcal{T}}^1} \hat{f}_e^t(d_e)
	+ \sum_{\forall (v,l,m) \in \mathcal{F}_{\!\mathcal{T}}^2} p_{v,l}^m d_v \label{lp:crit} \\[1mm]
	& \quad \text{subject to:}\;\text{\eqref{prec}*, \eqref{precend}*, \eqref{durstat}*, \eqref{durdyn}*, \eqref{timelags}*} \notag \\
	& \quad s_{a_2}+ n \mathrm{CT} \geq s_{a_1}+d_{a_1} \qquad \forall (a_1, a_2, n) \in \mathcal{D}_{\geq} \label{colres1} \\
	& \quad s_{a_2} + d_{a_2} + n \mathrm{CT} \leq s_{a_1} \qquad \forall (a_1, a_2, n) \in \mathcal{D}_{\leq} \label{colres2}
\end{alignat}

All the aforementioned constraints are always present in the above formulation; however,
these alone may not ensure feasibility because some collisions could occur.
For this reason, additional constraints \eqref{colres1} and/or \eqref{colres2}
could be iteratively added to heuristically resolve active collisions.
Thus, if a solution is feasible with respect to the current constraints of the reduced LP problem
but is not feasible for the original problem due to some active collisions,
then the worst collision (formally defined later) is detected and resolved by adding constraint \eqref{colres1} or \eqref{colres2}.
Afterward, the problem is resolved, and the same procedure is repeated if collisions still occur (see Figure \ref{fig:heur}).

To introduce the collision resolution more formally, constraints \eqref{colres1} and \eqref{colres2}
have to be shown as specialized forms of constraints \eqref{collisions1} and \eqref{collisions2}, respectively.
First, the variables $u_{a_i}^{g_i}$ in constraints \eqref{collisions1} and \eqref{collisions2} are fixed for tuple $\mathcal{T}$,
i.e., $u_{a_i}^{g_i} = 1$ if $(a_i, g_i) \in \mathcal{F}_{\!\mathcal{T}}^1$ or $(a_i, g_i, m_i) \in \mathcal{F}_{\!\mathcal{T}}^2$; otherwise, $u_{a_i}^{g_i} = 0$.
Second, it is sufficient to consider set $C_{\mathcal{T}} = \{\forall (a_i, a_j) : u_{a_i}^{g_i}+u_{a_j}^{g_j}=2, (a_i, g_i, a_j, g_j) \in C, a_i, a_j \in A_{\mathcal{T}}\}$
instead of $C$ because constraints \eqref{collisions1} and \eqref{collisions2} are neither binding nor violated for unselected locations and movements of tuple $\mathcal{T}$,
and each activity $a_i \in A_{\mathcal{T}}$ has an assigned movement or location.
Note that if $u_{a_i}^{g_i}+u_{a_j}^{g_j} = 2$, then the variable $c_{o}^n$ makes only one constraint, either \eqref{collisions1} or \eqref{collisions2}, active (``big M method'')
depending on the decision whether $s_{a_2}+n\mathrm{CT} \geq s_{a_1}+d_{a_1}$ or $s_{a_1} \geq s_{a_2}+d_{a_2}+n\mathrm{CT}$, which guarantees collision-free ordering.
However, these decisions correspond exactly to constraints \eqref{colres1} and \eqref{colres2}, in which the decision on ordering is given by adding
a triple $(a_i, a_j, n)$ to the related set $\mathcal{D}_{\geq}$ or $\mathcal{D}_{\leq}$, respectively; i.e., a constraint is added to the problem.
The constraint added is not removed during the solving of the reduced LP problem; therefore, this approach is heuristic.
The key question is how to select which constraint to add and how such constraint is related to the worst collision.
To find an answer, the maximal violation $\Gamma$ of constraints \eqref{collisions1} and \eqref{collisions2} for tuple $\mathcal{T}$ has to be defined as follows.

\begin{equation}
	\upsilon_{a_i, a_j}^n = s_{a_i}+d_{a_i}-s_{a_j}-n\mathrm{CT}
	\label{eq:first_res}
\end{equation}
\begin{equation}
	\mu_{a_i, a_j}^n = s_{a_j}+d_{a_j}+n\mathrm{CT}-s_{a_i}
	\label{eq:second_res}
\end{equation}
\begin{equation}
	\Gamma = \max_{\substack{\forall (a_i, a_j) \in C_{\mathcal{T}}\\\mathclap{\forall n \in \{-|\mathcal{R}|, \cdots, |\mathcal{R}|\}}}} \min (\upsilon_{a_i, a_j}^n, \mu_{a_i, a_j}^n)
	\label{eq:max_vio}
\end{equation}

Note that if $\Gamma \leq 0$, then there are no active collisions, and no extra constraints are needed;
failing that, i.e., $\Gamma > 0$, means that there is a pair of colliding activities that is not time disjunctive.
In that case, let $(a_i^*, a_j^*, n^*)$ be an optimal argument of the max function in Equation \eqref{eq:max_vio} (replace $\max$ with $\text{argmax}$).
This triple defines the worst collision, i.e., the collision with the biggest time intersection, occurring in a current solution.
Whether this collision should be resolved by adding a constraint of type \eqref{colres1} or \eqref{colres2}
is determined by the $\upsilon^* = \upsilon_{a_i^*, a_j^*}^{n^*}$ and $\mu^* = \mu_{a_i^*, a_j^*}^{n^*}$ values.
If $\upsilon^* \leq \mu^*$ ($\mu^* \leq \upsilon^*$), then
the worst collision is resolved by adding the triple to $\mathcal{D}_{\geq}$ ($\mathcal{D}_{\leq}$) because it may result in
smaller changes in timing after the problem is resolved with the added constraints.

\subsection{Sub-heuristics}\label{sec:subheur}


The aim of the sub-heuristics is to modify a given tuple $\mathcal{T}$ and its timing calculated by the reduced LP problem
such that the energy consumption would be reduced in successive LP calls.
Because the performed modifications can result in a violation of time lags or the occurrence of collisions,
a penalty based on the duration of breakage and average input power is added to these modifications.
If the modifications of an active sub-heuristic do not lead to significant energy improvements,
then the next sub-heuristic is selected in round-robin order.

The goal of the \emph{(de)select power mode} sub-heuristic, which focuses on the application of the power-saving modes of robots,
is to find and apply an alternative power-saving mode for some activity $v \in V$.
To select a suitable mode and activity, the energy consumption
is estimated for all applicable modes of each $v \in V$ as follows.
If a different mode $m \in M_r$ of robot $r \in \mathcal{R}$ is applied for activity $v'$,
then some activities $v \in V_r$ are uniformly shortened/prolonged to meet the cycle time.
After the timing is modified, both the energy consumption and the above-mentioned penalty are determined.
Finally, the choice falls on mode $m$ and activity $v'$ with the lowest sum of the energy consumption and penalty.
Note that the sub-heuristic uses a Tabu list, i.e., a short-term memory with forbidden modifications,
to avoid cycling; therefore, some power-saving modes may not be applicable for some activities.

The \emph{change locations} sub-heuristic optimizes the closed paths of robots, i.e.,
$\forall \mathcal{HC}_r^{\mathrm{loc}} \in \mathscr{P}$, by modifying the go-through locations as follows.
Let `$\dashrightarrow l_a \xrightarrow{t_1} l_b \xrightarrow{t_2} l_c \dashrightarrow$'
be a part of $\mathcal{HC}_r^{\mathrm{loc}}$, where $t_1$ and $t_2$ are the movements between the locations;
then, the question arises as to whether the inner part, i.e., `$\xrightarrow{t_1} l_b \xrightarrow{t_2}$', can
be replaced so as to achieve a reduction in energy consumption.
To find an answer, each viable substitution `${\color{gray} l_a} \xrightarrow{t_{\vdash}} l_{\circ} \xrightarrow{t_{\dashv}} {\color{gray} l_c}$'
has to be evaluated in terms of energy by solving the convex problem \eqref{gs:crit} to \eqref{gs:constr},
where $t_1, t_{\vdash} \in T_{e_1}$, $t_2, t_{\dashv} \in T_{e_2}$, $e_1, e_2 \in E_\mathcal{T}$, $l_b, l_{\circ} \in L_{v_b}$, $v_b \in V$,
and $K = d_{e_1}^{\mathrm{LP}}+d_{e_2}^{\mathrm{LP}}$ and $d_{v_b}^{\mathrm{LP}}$ are constants determined from the LP solution.
Because the problem is convex and one-dimensional, the golden section search algorithm can be used to find the optimal solution.
Then, the most energy-friendly substitution not breaking the spatial compatibility is applied, and the process is repeated for other sub-paths.

\begin{alignat}{1}
	& \text{minimize}\;f_{e_1}^{t_{\vdash}}(d_{e_1})+f_{e_2}^{t_{\dashv}}(K-d_{e_1})
	+ p_{v_b,l_{\circ}}^m d_{v_b}^{\mathrm{LP}} \label{gs:crit} \\[1mm]
	& \qquad \max (\underline{d}_{e_1}^{t_{\vdash}}, K-\overline{d}_{e_2}^{t_{\dashv}}) \leq d_{e_1}
	\leq \min (\overline{d}_{e_1}^{t_{\vdash}}, K-\underline{d}_{e_2}^{t_{\dashv}}) \label{gs:constr}
\end{alignat}

The last sub-heuristic, called \emph{change path}, diversify the search process to
allow exploring some otherwise unreachable $\mathcal{HC}_r^{loc}$.
The sub-heuristic selects one $\mathcal{HC}_r^{loc} \in \mathscr{P}$ and randomly changes its
go-through locations such that spatial compatibility is achieved and the resulting
closed path exists.

\section{Experimental Results}\label{sec:res}

To evaluate the effectiveness and performance of the proposed approaches,
the algorithms were tested on benchmark instances with 5, 8, and 12 robots.
The experiments were carried out on a Linux server with two Intel Xeon E5-2620 v2
2.10GHz processors (2\,x\,6 physical cores + hyper-threading) and 64 GB of DDR3 memory,
in which Gurobi 6.0.4, a state-of-the-art MILP solver, and lp\_solve 5.5.2.0,
an open-source MILP solver, were installed to solve LP and MILP problems.
The optimization program, written in C++11 and compiled by GCC 4.9.2, was
benchmarked on three generated data sets: S5 (small, 5 robots), M8 (medium, 8 robots),
and L12 (large, 12 robots), each of which contains 10 instances of the problem.
For the purposes of all experiments, each energy function $f_e^t$
was approximated by 10 linear functions, i.e., $|B| = 10$, and
the minimal number of optimization iterations per tuple $\Phi_{\mathrm{max}}$
was set to 100, 600, and 1000 for the S5, M8, and L12 data sets, respectively.
The data sets, their configuration files, and the generator are publicly available
on \IIGsources to enable comparisons with other research.

\begin{table}
	\setlength\dashlinedash{0.2pt}
	\setlength\dashlinegap{1.5pt}
	\setlength\arrayrulewidth{0.3pt}
	\caption{Performance of LP solvers for the heuristic.}
	\centering
	\begin{tabular}{lc|ccc}
		& & lp\_solve & Gurobi & Gurobi CF \\ \hline
		\multirow{2}{0.3cm}{S5} & sequential & 23.1 & 78.5 & 365 \\
		& parallel & 304 & 1111 & 4708 \\ \hdashline
		\multirow{2}{0.3cm}{M8} & sequential & 7.9 & 44.2 & 170 \\
		& parallel & 93.4 & 488 & 2134 \\ \hdashline
		\multirow{2}{0.3cm}{L12} & sequential & 3.3 & 30.8 & 97.9 \\
		& parallel & 42.6 & 371 & 1084 \\ \hline
	\end{tabular}
	\label{tab:perf}
\end{table}

\begin{figure}[!t]
	\centering
	\includegraphics[width=0.5\textwidth]{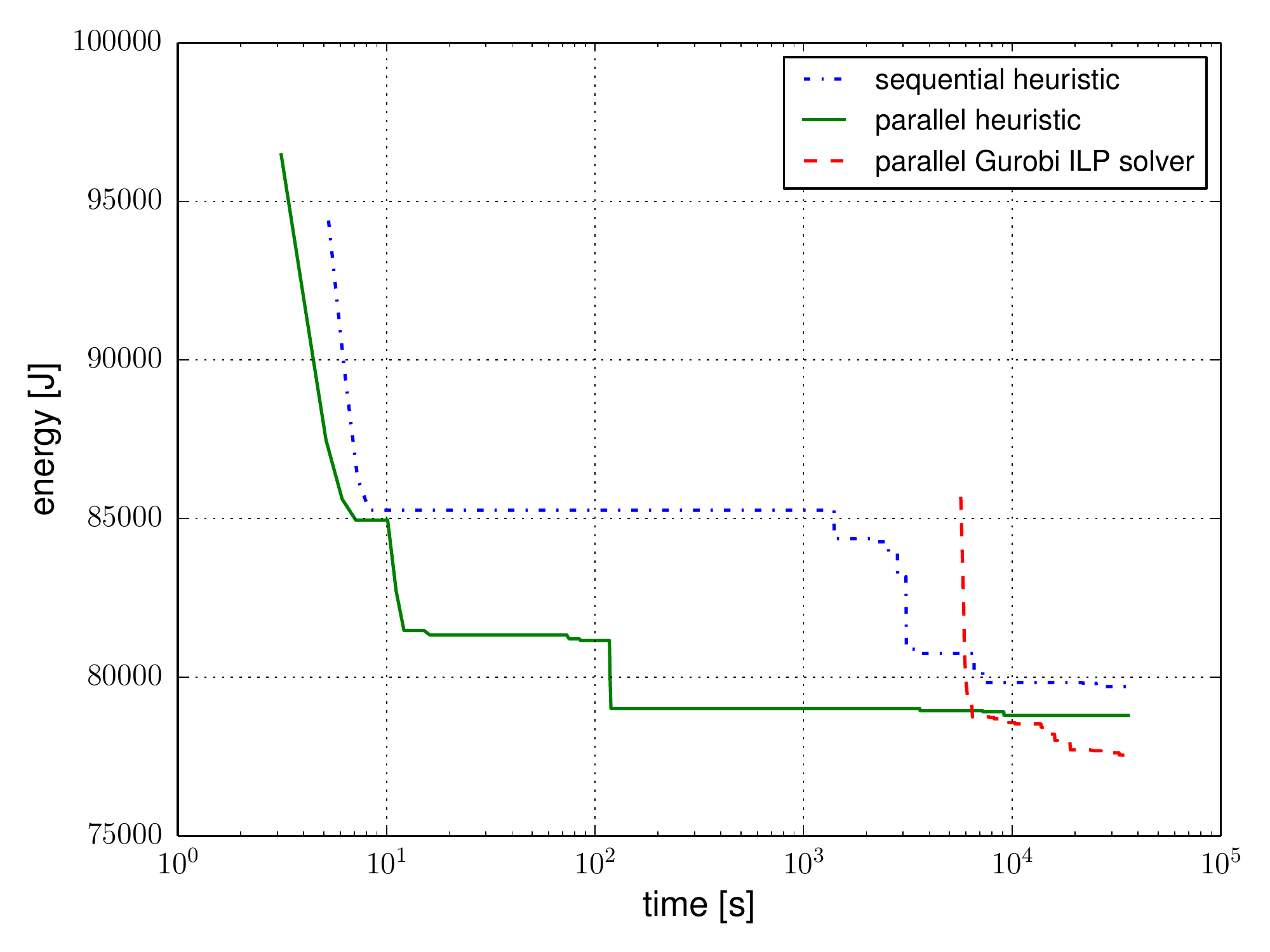}
	\caption{Progress of the heuristic and MILP solver on M8\_8 instance.}
	\label{fig:conv}
\end{figure}

\begin{table*}
	\caption{Quality of solutions in terms of energy consumption for the S5 data set.}
	\centering
	\begin{tabular}{c||ccc|c||ccc|c||c}
		\multicolumn{1}{c||}{} & \multicolumn{3}{c}{heuristic ($t_{max}$ = 30\,s)} & \multicolumn{1}{c||}{MILP ($t_{max}$ = 30\,s)} &
		\multicolumn{3}{c}{heuristic ($t_{max}$ = 1\,h)} & \multicolumn{1}{c||}{MILP ($t_{max}$ = 1\,h)} & \multicolumn{1}{c}{} \\ \hline
		instance & best run & avg run & worst run & avg run & best run & avg run & worst run & avg run & \multicolumn{1}{c}{lower bound} \\ \hline
		S5\_0 & 40654.80 & 40788.45 & 40936.20 & 42521.9 & 40494.90 & 40637.85 & 40776.60 & 40272.8 & 39849.4 \\
		S5\_1 & 31195.10 & 31335.18 & 31479.20 & 34582.6 & 31052.90 & 31111.81 & 31184.60 & 30623.3 & 30208.2 \\
		S5\_3 & 47482.80 & 47665.48 & 47861.60 & -- & 47434.10 & 47483.28 & 47674.90 & -- & 46374.3 \\
		S5\_4 & 47826.30 & 48039.38 & 48326.60 & -- & 47541.10 & 47739.06 & 48067.60 & -- & 45922.5 \\
		S5\_5 & 41692.00 & 41902.74 & 42081.20 & -- & 41654.70 & 41699.96 & 41741.00 & 41498.8 & 40349.3 \\
		S5\_6 & 34920.30 & 35043.71 & 35128.20 & 36315.6 & 34760.00 & 34856.44 & 34918.90 & 34720.9 & 34398.7 \\
		S5\_7 & 42511.60 & 42821.70 & 42987.60 & 45670.2 & 42384.40 & 42515.82 & 42627.60 & 42033.0 & 40801.4 \\
		S5\_8 & 39181.00 & 39423.62 & 39602.40 & 41440.2 & 39174.00 & 39301.44 & 39464.10 & 38364.8 & 37772.9 \\
		S5\_9 & 38864.90 & 39067.57 & 39235.90 & 39721.5 & 38649.80 & 38840.31 & 38924.80 & 38283.8 & 37923.5 \\ \hline
	\end{tabular}
	\label{tab:qualityS5}
\end{table*}

\begin{table*}
	\caption{Quality of solutions in terms of energy consumption for the M8 data set.}
	\centering
	\begin{tabular}{c||ccc|c||ccc|c||c}
		\multicolumn{1}{c||}{} & \multicolumn{3}{c}{heuristic ($t_{max}$ = 600\,s)} & \multicolumn{1}{c||}{MILP ($t_{max}$ = 600\,s)} &
		\multicolumn{3}{c}{heuristic ($t_{max}$ = 1\,h)} & \multicolumn{1}{c||}{MILP ($t_{max}$ = 1\,h)} & \multicolumn{1}{c}{} \\ \hline
		instance & best run & avg run & worst run & avg run & best run & avg run & worst run & avg run & \multicolumn{1}{c}{lower bound} \\ \hline
		M8\_0 & 86831.60 & 90580.66 & 95333.80 & -- & 87365.40 & 90258.81 & 94670.50 & -- & 78641.3 \\
		M8\_1 & 88098.90 & 88491.50 & 88883.80 & 89297.4 & 87999.70 & 88236.49 & 88488.00 & 89182.7 & 83677.7 \\
		M8\_2 & 90048.70 & 90667.98 & 91463.60 & 92991.0 & 89939.70 & 90600.33 & 91804.20 & 89744.0 & 82503.9 \\
		M8\_3 & 83412.60 & 83948.53 & 84942.60 & -- & 83224.70 & 83709.86 & 84741.30 & 83621.6 & 78564.3 \\
		M8\_4 & 76914.70 & 77549.80 & 78191.60 & 82881.8 & 77058.90 & 77400.65 & 77723.00 & 76582.8 & 70838.1 \\
		M8\_5 & 89239.40 & 90337.15 & 91486.00 & -- & 89369.40 & 89873.88 & 90545.30 & -- & 81856.3 \\
		M8\_6 & 95314.50 & 96021.27 & 96629.40 & 97875.4 & 95638.50 & 95836.32 & 96134.60 & 94918.1 & 88023.1 \\
		M8\_7 & 83777.20 & 85218.11 & 86951.50 & -- & 83016.50 & 84881.71 & 86266.50 & -- & 77038.8 \\
		M8\_8 & 78514.40 & 79220.31 & 80530.70 & -- & 78465.80 & 78859.41 & 79342.40 & -- & 74447.2 \\
		M8\_9 & 92875.60 & 93309.59 & 93736.40 & 93951.9 & 91960.60 & 92942.01 & 93654.10 & 92521.9 & 86900.0 \\ \hline
	\end{tabular}
	\label{tab:qualityM8}
\end{table*}

\begin{table}
	\caption{Quality of solutions for the L12 data set.}
	\centering
	\begin{tabular}{c||c||c}
		\multicolumn{1}{c||}{} & \multicolumn{1}{c||}{heuristic ($t_{max}$ = 3\,h)} & \multicolumn{1}{c}{} \\ \hline
		instance & 3 runs & \multicolumn{1}{c}{lower bound} \\ \hline
		L12\_0 & --, --, \textbf{204464} & 167229 \\
		L12\_1 & --, \textbf{177731}, -- & 154744 \\
		L12\_2 & \textbf{211019}, --, 211377 & 173457 \\
		L12\_3 & --, --, \textbf{188527} & 155121 \\
		L12\_4 & 187163, --, \textbf{171860} & 148441 \\
		L12\_5 & --, --, -- & -- \\
		L12\_6 & --, --, --  & -- \\
		L12\_7 & \textbf{190060}, 202737, -- & 155598 \\
		L12\_8 & \textbf{218844}, --, -- & 176156 \\
		L12\_9 & \textbf{173743}, 176969, 176567 & 152531 \\ \hline
	\end{tabular}
	\label{tab:qualityL12}
\end{table}

The first experiment measures the effect of the parallelization
and Gurobi simplex method on the performance of the hybrid heuristic.
A good indicator of the performance is the number of LP evaluations per second
because more than 90\,\% of the computational time is consumed by LP evaluations, i.e.,
the building of the LP problem, its optimization, and the extraction of a solution.
Table~\ref{tab:perf} shows the average number of LP evaluations per second for each data set.
The figures indicate that the parallel heuristic is about 12 times faster than the sequential one,
and the specialized Gurobi simplex method, denoted as `Gurobi CF',
accelerates the heuristic about 3 to 4 times compared with the regular one;
therefore, an overall speedup of about 36 to 48 can be expected for the aforementioned configuration.
For comparison, Table~\ref{tab:perf} also presents the corresponding values for the lp\_solve open-source solver.
To show that a higher number of evaluated tuples also has a positive impact on the quality of solutions,
the dependence of the criterion value, i.e., energy consumption, on the time limit was plotted in Figure~\ref{fig:conv}.
The results on an instance with 8\,robots revealed that the parallel heuristic with 24 threads (12 cores + hyper-threading)
converged significantly faster than the sequential version; therefore, a similar solution quality was achievable in a fraction of the time.
In comparison to the Gurobi MILP solver, the heuristic seems to be stronger in finding feasible solutions,
compare 1.5\,h with 3.1\,s (5.3\,s) required by the parallel (sequential) heuristic, and is more suitable for a short-term (re)optimization.
On the other hand, if the MILP solver is given enough time, then better solutions may be found for the medium instances (see Figure~\ref{fig:conv}).
The same experiment was repeated on L12\_9 instance with 12 robots.
The parallel (sequential) heuristic found the first feasible solution in 10\,m (4\,h) and the best criterion was 177276\,J (186855\,J) after the 10-h time limit.
The Gurobi MILP solver had run out of memory (installed 64\,GB of memory) in less than 4 hours without having any feasible solution.

The second experiment evaluates the quality of solutions in terms of
energy consumption (lower energy consumption is better)
for both the parallel heuristic and the parallel Gurobi MILP solver.
The quality of obtained solutions is compared with a tight lower bound,
calculated as the sum of the lower bounds on energy consumption of individual robots
in which global constraints are neglected (see Equations \eqref{ilp:criterion} to \eqref{durdyn}),
to obtain an upper estimate of the gap from optimal solutions.
The run time for the bound calculation was limited to 5 hours.
To ensure that the measured data are statistically significant, the best,
average, and worst qualities of solutions are determined from 10 runs for data sets S5 and M8.
Because the biggest data set (L12) is too computationally demanding,
only 3 measurements were carried out for each instance,
and the criterion values were stated explicitly.
Surprisingly, the Gurobi MILP solver has a deterministic behavior
and thus provided the same quality of solutions for all runs.
That is why only the average is stated for the Gurobi MILP solver.

Tables~\ref{tab:qualityS5} to \ref{tab:qualityL12} show the qualities of solutions for the S5, M8, and L12 data sets, respectively.
Instance S5\_2 is omitted from the results because it was proved to be infeasible.
For small instances, the best solutions found by the heuristic and the MILP solver
were near the optimal ones because the average gap from the lower bound was about 2\,\%.
On one hand, the Gurobi MILP solver could find almost optimal solutions for some small instances,
on the other hand, it had difficulty in providing feasible solutions even for instances with 5\,robots.
Compared with the MILP solver, the heuristic obtained very good solutions in a short time,
and its ability to find feasible solutions seemed to be significantly better.
The results for medium instances indicated a similar trend.
The heuristic solved all the instances, whereas the MILP solver found a solution for only 6 instances in the 1-h time limit.
Moreover, the quality of solutions achieved by the heuristic in the 1-h time limit was comparable with that of the MILP solver,
and the average gap increased to about 7.5\,\% for the best solutions found.
The largest instances were intractable for the MILP solver;
therefore, only the results for the heuristic are presented.
The heuristic solved 8 of 10 instances in 3 runs for the time limit of 3 hours.
Two instances remained unsolved because they were either too difficult to solve or infeasible.

The last experiment shows how the robot cycle time, which was scaled by 1.0, 1.1, and 1.2 factors,
respectively, influences the performance of the heuristic and MILP solver.
The summary of results for the S5 data set is in Table~\ref{tab:ct_analysis},
where each figure is the average criterion value from 10 runs with $t_{max}$ = 600\,s.
The outcomes indicate that the heuristic outperforms the MILP solver if the cycle time is tight,
i.e., a feasible solution is hard to find due to the limited time for the movements and operations.
However, if the cycle time is prolonged, then the MILP solver gets ahead because
its ability to find optimal solutions by a systematic search becomes dominant for the given time limit.

\begin{table}
	\caption{Dependence of the quality of solutions on the cycle time.}
	\centering
	\begin{tabular}{|c|c|c|c|c|c|c|} \cline{2-7}
		\multicolumn{1}{c|}{\rule{0pt}{8pt}} & \multicolumn{2}{c|}{$\mathrm{CT}$} & \multicolumn{2}{c|}{$\mathrm{1.1*CT}$} & \multicolumn{2}{c|}{$\mathrm{1.2*CT}$} \\ \hline
		instance & heur. & MILP & heur. & MILP & heur. & MILP \\ \hline
		S5\_0 & 40717 & \textbf{40355} & 40435 & \textbf{40022} & 41381 & \textbf{40895} \\
		S5\_1 & 31198 & \textbf{30812} & 29410 & \textbf{29269} & 29433 & \textbf{29239} \\
		S5\_3 & \textbf{47579} & -- & \textbf{44177} & 44338 & 43846 & \textbf{43284} \\
		S5\_4 & \textbf{47749} & -- & \textbf{43993} & 44233 & 43577 & \textbf{43400} \\
		S5\_5 & \textbf{41730} & -- & 38248 & \textbf{38227} & 37644 & \textbf{37566} \\
		S5\_6 & \textbf{34880} & 34980 & 32359 & \textbf{32186} & 31744 & \textbf{31630} \\
		S5\_7 & \textbf{42560} & 42619 & 42230 & \textbf{41867} & 42737 & \textbf{42571} \\
		S5\_8 & 39360 & \textbf{38761} & 38407 & \textbf{38045} & 39051 & \textbf{38432} \\
		S5\_9 & 38935 & \textbf{38412} & 38769 & \textbf{38233} & 39695 & \textbf{39302} \\ \hline
	\end{tabular}
	\label{tab:ct_analysis}
\end{table}

The approach proposed in this study cannot be directly compared with existing works \cite{OptRob1, OptRob6, OptRob7}
because the robot cycle time is considered instead of the work cycle time.
However, in general, much bigger instances were solved compared with the aforementioned works, which considered only one to four robots per robotic cell.
Besides, the proposed approach took into account additional optimization aspects, such as the robot power-saving modes and positions.
Finally, the algorithms were tested on more problem instances than those in others works.
The data sets are available for others.

\section{Case Study from \v{S}koda Auto}\label{sec:casestudy}







This case study shows a potential impact of the optimization on the energy consumption
of existing robotic cells by considering a long-operating robotic cell from \v{S}koda Auto
(see Figure \ref{fig:rob_cell} for a screenshot from the simulation), in which a part of
an automotive body is welded, glued, and assembled by 6 industrial robots with a robot cycle time of 56 seconds.
The timing of individual robotic operations was obtained from robotic programs.
An alternative way is to measure and subsequently identify the movements.
The optimized speed of movements can be easily entered into the robotic programs.
The energy function $f_e^t$ of a trajectory was fitted from the points
obtained from simulations in Siemens Tecnomatix Process Simulate (Digital Factory - Industry 4.0),
in which the robot controller supported the calculation of the energy consumption
of movements according to the Realistic Robot Simulation standard.
It is also possible to carry out the measurements with a physical robot
to obtain the energy functions; however, this is impractical in existing robotic cells.
Alternative approaches to the modeling and simulation are found in \cite{altsimmod1,altsimmod2}.
To ensure the repeatability of the production process in terms of output quality,
the welding, gluing, and assembling operations remained the same,
i.e., the fixed duration and energy consumption were extracted from the measured power profiles.
Only the robot speeds and power saving-modes (at home position) were addressed in the optimization
because minimal intervention is desirable for existing robotic cells.
More information about the structure of the robotic cell, the timing, and the synchronizations can be obtained from a published instance file.

Based on the results, it was estimated that the original energy consumption of 500\,kJ (maximal speeds, without power-saving modes)
could be decreased to 391\,kJ (reduced speeds, with power-saving modes) per cycle, resulting in about 20\,\% energy savings.
The power-saving modes of robots saved about 2.4\,\% of energy, whereas the remaining savings were attributed to the optimization of speeds.
If the cycle time is extended to 70\,s and 80\,s, then the application of energy-saving modes would improve the consumption
by about 6.8\,\% and 12\,\%, respectively, compared with their nonuse.
This finding  may be particularly useful during over production or production cuts.
Note that the inclusion of the bus power-off mode is not always straightforward
because it requires interaction between the robot and a superior controller, which may not be ready in existing cells.
However, the effort to implement such interaction is not high either.
The outcomes of the optimization indicate that a significant reduction in energy consumption can be achieved for existing robotic cells
and that even more can be expected for planned robotic cells that will maximize the potential of the optimization algorithm.

\section{Conclusion}\label{sec:conclusions}




Energy optimization is undoubtedly a current and important problem for the industry
because such optimization could lead to a significant decrease in costs.
Besides being able to save money, an involved company could also improve
its green credentials and become more competitive.

This work presents a holistic approach to the energy optimization
of robotic cells that considers many optimization aspects.
A universal mathematical model for describing robotic cells is proposed,
from which is derived an MILP formulation that is directly solvable by MILP solvers.
However, these solvers can solve only small instances.
Therefore, a hybrid parallel heuristic is devised for instances with up to 12 robots.
The strength of the heuristic is that its performance scales almost linearly up to 12 cores;
thus, significant acceleration is achievable on modern processors.
Moreover, the heuristic uses a specialized Gurobi simplex method for piecewise linear
convex functions that is about 3 or 4 times faster than broadly used simplex methods.
The merits of the heuristic are verified by experiments, the results of which showed
that the heuristic found a solution to 27 of 29 problems (+1 infeasible)
and clearly outperformed the MILP solver for the largest problems.
Note that all the algorithms, the generator, and the generated data sets are publicly available from \IIGsources;
thus, other researchers can experiment with the heuristic and explore the documented open-source code.

Finally, a real robotic cell from \v{S}koda Auto is optimized.
The results indicate that energy savings of up to 20\,\% can be achieved
merely by changing the robot speeds and applying power-saving modes.
In future studies, we plan to fully optimize another robotic cell
and integrate our approach into industrial software.

\section*{Acknowledgment}

This work was supported by the Grant Agency of the Czech Republic under the Project FOREST GACR P103-16-23509S.

The authors are also very grateful to Antonin Novak from the Czech Technical University in Prague
for his advice of using special Gurobi simplex method.

\ifCLASSOPTIONcaptionsoff
  \newpage
\fi

\bibliographystyle{IEEEtran}
\bibliography{references}

\begin{IEEEbiography}[{\includegraphics[width=1in,height=1.25in,clip,keepaspectratio]{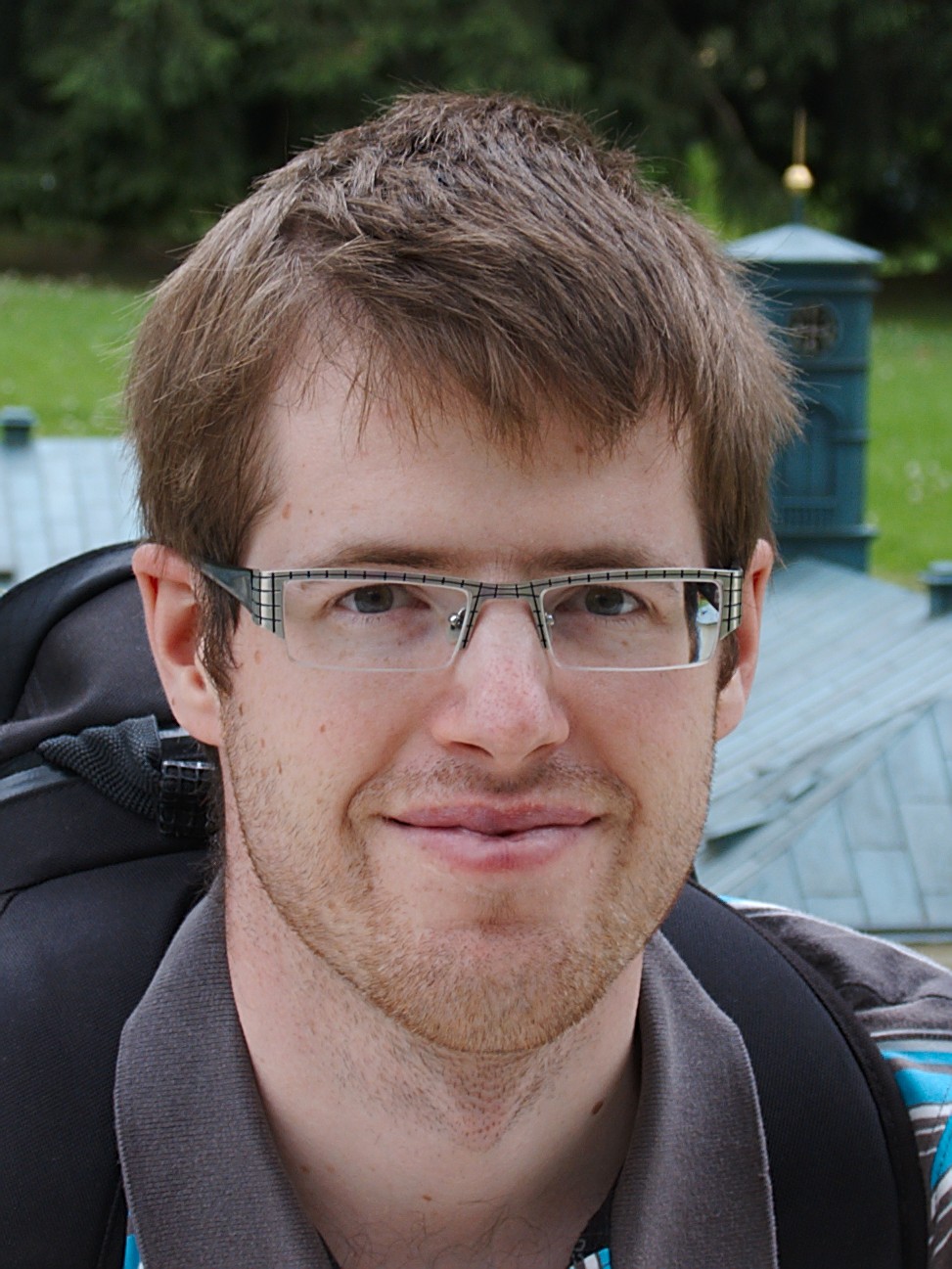}}]{Libor Bukata}
received his Master's degree in Software Engineering from the Czech Technical University in Prague (CTU) in 2012,
where he currently conducts the research on the energy optimization of robotic cells as a post-graduate student.
Up to now, he has published 1 journal paper (Journal of Parallel and Distributed Computing)
and more than 3 contributions to international conferences (PDP 2013, SCOR 2014, ECCO 2014, MISTA 2015).
Moreover, he reviewed some papers from MISTA 2015 conference, where he also participated in organization.
His research reflects his passion for scheduling, combinatorial optimization, and parallel and distributed systems.
\end{IEEEbiography}

\begin{IEEEbiography}[{\includegraphics[width=1in,height=1.25in,clip,keepaspectratio]{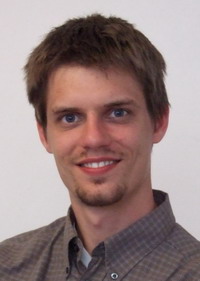}}]{P\v{r}emysl \v{S}\r{u}cha}
graduated in Technical Cybernetics from the Czech Technical University in Prague (CTU) in 2003
and received a PhD degree in Electrical Engineering and Informatics from CTU in 2007.
From 2011 to 2012 he was in a postdoc position at LAAS-CNRS, Toulouse, France.
Since 2012 he works as an Assistant Professor at CTU.
His main research activities are in the area of scheduling algorithms and parallel programming.
He participated in several industrial projects (with e.g. Skoda, Air Navigation Services of the Czech Republic)
and he was a national coordinator in the Design, Monitoring and Operation of Adaptive Networked Embedded Systems project (DEMANES, ARTEMIS Joint Undertaking).
He published more than 40 papers in international journals and conferences; he was a co-chair of the Multidisciplinary International Scheduling Conference (MISTA 2015).
\end{IEEEbiography}

\begin{IEEEbiography}[{\includegraphics[width=1in,height=1.25in,clip,keepaspectratio]{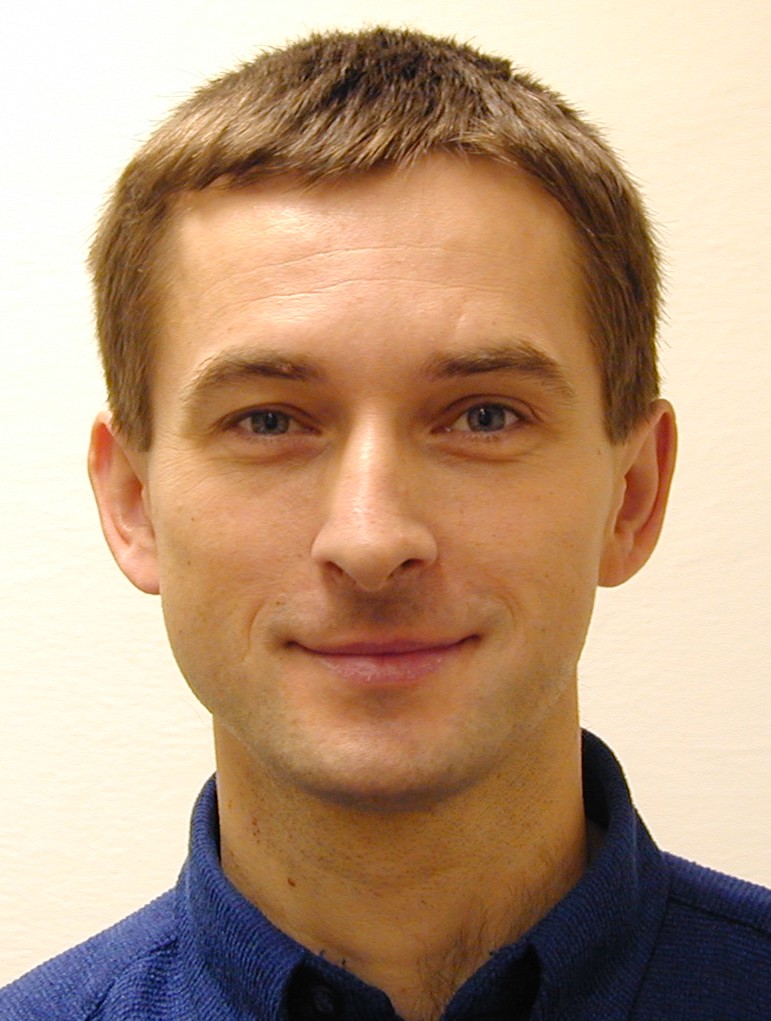}}]{Zden\v{e}k Hanz\'{a}lek}
graduated in Electrical Engineering at the Czech Technical University (CTU) in Prague in 1990.
He obtained his Ph.D. degree in Industrial Informatics from the Universite Paul Sabatier Toulouse and Ph.D.
degree in Control Engineering from the CTU.
He worked on optimization of parallel algorithms at LAAS CNRS in Toulouse
and on discrete event dynamic systems at LAG INPG in Grenoble.
He founded and coordinated the Industrial Informatics Group at CTU Prague focusing on scheduling,
combinatorial optimization algorithms, real-time control systems and industrial communication protocols.
From 2011 to 2014, he served as Senior Manager of newly established Mechatronics group at Porsche Engineering Services in Prague.
He has been involved in many industrial contracts (e.g. Skoda, Porsche, Volkswagen, Rockwell, Air Navigation Services, EATON).
He is currently working on different aspects of scheduling and optimization problems used in production and in time-triggered communication protocols.
\end{IEEEbiography}

\begin{IEEEbiography}[{\includegraphics[width=1in,height=1.25in,clip,keepaspectratio]{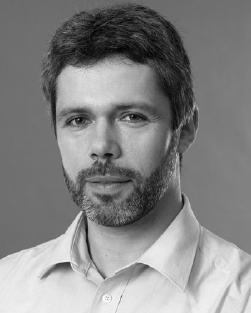}}]{Pavel Burget}
received the M.Sc. degree in computer engineering at Czech Technical University (CTU) in Prague, Czech Republic in 1996.
In 2008 he received his Ph.D. degree in Control Engineering and Robotics at the CTU in Prague.
He is assistant professor at the Department of Control Engineering at CTU in Prague.
He has been taking teaching and research responsibilities in the areas of industrial communications and control systems.
He has had several publications in journals (e.g. IEEE T on Industrial Informatics, Control Engineering Practice)
and conferences focusing on industrial control systems (e.g. ETFA, CASE, ECRTS, IFAC World Congress).
\end{IEEEbiography}

\end{document}